\documentclass[11pt]{article}

\usepackage[final]{acl}

\usepackage{times}
\usepackage{latexsym}

\usepackage[T1]{fontenc}

\usepackage[utf8]{inputenc}

\usepackage{microtype}

\usepackage{inconsolata}

\usepackage{graphicx}

\usepackage{amsmath}
\usepackage{amssymb}
\usepackage{booktabs}
\usepackage{multirow}
\usepackage{makecell}
\usepackage{adjustbox}

\usepackage[most]{tcolorbox}
\usepackage{listings}
\usepackage{xcolor}
\usepackage[table]{xcolor}
\definecolor{methodgreen}{RGB}{230,245,230}

\definecolor{promptbackground}{RGB}{248,248,248}
\definecolor{promptborder}{RGB}{170,170,170}
\definecolor{prompttitle}{RGB}{240,240,240}

\newtcblisting{promptbox}[1]{
    enhanced,
    breakable,
    listing only,
    title={#1},
    fonttitle=\bfseries,
    coltitle=black,
    colbacktitle=prompttitle,
    colback=promptbackground,
    colframe=promptborder,
    boxrule=0.4pt,
    arc=1mm,
    outer arc=1mm,
    left=1.5mm,
    right=1.5mm,
    top=1mm,
    bottom=1mm,
    toptitle=0.8mm,
    bottomtitle=0.8mm,
    before skip=6pt,
    after skip=8pt,
    listing options={
        basicstyle=\ttfamily\footnotesize,
        breaklines=true,
        breakatwhitespace=false,
        columns=fullflexible,
        keepspaces=true,
        showstringspaces=false,
        tabsize=2
    }
}

\title{TabScope: Question-Adaptive Scope Selection for Table Question Answering}

\author{
 {
  \textbf{Yuxiang Wang}, 
  \textbf{Junhao Gan}, 
  \textbf{Jianzhong Qi} 
  }
  \\
  {The University of Melbourne} 
  \\ \texttt{yuxiang.wang8@student.unimelb.edu.au}\\ 
   \texttt{\{junhao.gan, jianzhong.qi\}@unimelb.edu.au}
}

\begin{document}
\maketitle
\newcommand{\model}{\textsc{TabScope}}
\newcommand{\wdata}{\texttt{WikiTQ}}
\newcommand{\ldata}{\texttt{SLQA}}
\newcommand{\subdata}{\textsc{WTQ-SubTab}}

\begin{abstract}
Large Language Models (LLMs) have shown strong performance on table question answering, yet their accuracy often degrades as table size increases. We find that this degradation is not uniform across question types. Localization-sensitive questions are particularly affected by irrelevant table content, while questions requiring broader evidence may still benefit from full-table reasoning. Based on this observation, we propose a question-adaptive framework that dynamically selects between localized and full-table reasoning. The framework constructs question-specific sub-tables through operation-aware table decomposition and uses the predicted question type to determine the appropriate reasoning mode. We further introduce silver reference sub-tables for evaluating evidence selection and construct \ldata{}, a benchmark based on real-world long tables. Experiments on WikiTQ and \ldata{} show that localization is particularly effective for lookup and local reasoning questions, while adaptive selection between localized and full-table reasoning achieves the best overall performance. These results highlight that long-table QA requires deciding not only how to localize, but also when to localize. Our code and datasets will be made available upon publication of the paper.

\end{abstract}

\section{Introduction}
\label{sec:introduction}

Large Language Models (LLMs) have demonstrated strong reasoning
capabilities for table question answering (TableQA), particularly when combined with chain-of-thought (CoT) prompting
\citep{wei2022chain,Chen23, tablap, tabgr}. However, their performance remains sensitive to table size. As tables grow longer, LLMs struggle to
identify the small set of rows and columns needed to support the answer from noisy inputs, leading to lower answer accuracy
\citep{ye2023large,LiuLHPBPL24, tabsd, Enotab}.

\begin{figure}[!t]
    \centering
    \includegraphics[width=\columnwidth]{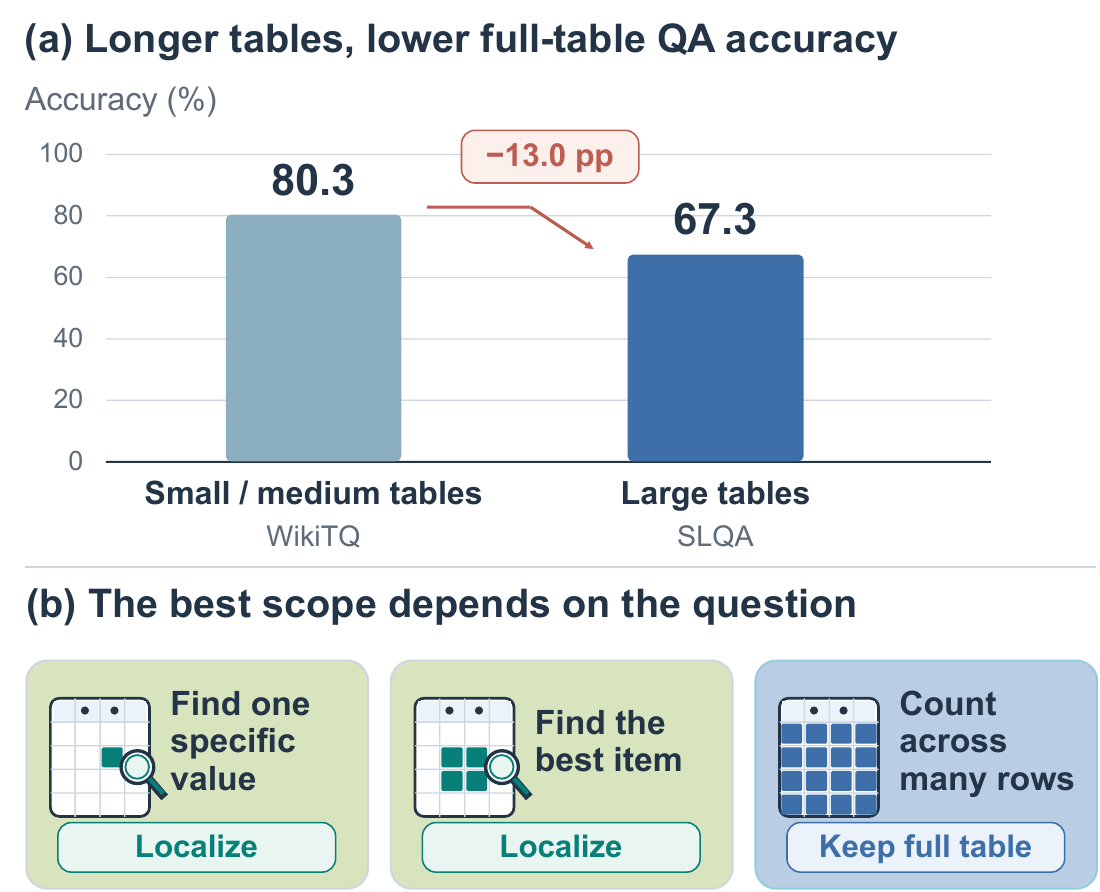}
   \caption{
Motivation for question-adaptive table scope selection. (a) Full-table CoT exact-match accuracy on \wdata{} and \ldata{}, averaged across GPT-5-mini and LLaMA-3.3-70B. The benchmarks differ in both table size and data distribution. (b) Different questions require different evidence scopes, motivating selection between localized and full-table reasoning.
}
    \label{fig:intro_motivation}
\end{figure}

Our analysis shows that this degradation varies across question types.
Localization is particularly effective when the answer depends on a
small, restricted table region. In contrast, questions requiring broad
comparison sets, complete aggregation domains, or information
distributed across many rows may benefit from retaining the full table.
Decomposition may otherwise remove necessary context or introduce
retrieval errors. Figure~\ref{fig:intro_motivation} summarizes these
observations. Full-table reasoning performs worse on large tables, while
the benefit of localization varies with both the question and table
size. This leads to the central question of our work: \emph{when should
a TableQA model localize the table, and when should it reason over the
full context?}

We regard TableQA as a question-adaptive reasoning problem rather than
a fixed decomposition pipeline, especially for long tables. We propose
\model{}, which first determines whether a question is better served by
the full table or a compact sub-table. When localization is expected to
be beneficial, \model{} uses an operation-aware decomposer with
refinement to select the relevant rows and columns before reasoning.
Otherwise, the model reasons directly over the full table. This design
uses compact table regions when helpful while preserving broader
context when necessary.

Existing TableQA benchmarks present two additional limitations for studying evidence localization. First, datasets such as WikiTableQuestions~\citep{pasupat} provide only tables, questions, and final answers, without explicit annotations of the supporting sub-tables. This makes it difficult to evaluate the quality of intermediate evidence selection. 
Second, most existing benchmarks~\cite{pasupat, ChenWCZWLZW20, tablebench} contain relatively small tables and provide limited coverage of long-table settings. We address the first limitation by constructing silver reference sub-tables for evaluating decomposition quality. For the second, we develop an automatic pipeline for generating question--answer pairs from real-world tables and use it to construct \ldata, a benchmark for long-table question answering.

Our main contributions are as follows:
\begin{itemize}
\item We construct silver reference sub-tables for directly evaluating decomposition quality and propose an operation-aware decomposition method that improves evidence selection.

\item We analyze when to localize and introduce a question-adaptive framework selects between localized and full-table reasoning.

\item We develop an automatic pipeline for generating validated question--answer pairs from real-world tables and use it to build \ldata{}, a benchmark for long-table question answering.

\end{itemize}

\section{Related Work}
\label{sec:related}
Recent LLM-based TableQA methods increasingly address the difficulty of reasoning over large tables by reducing, retrieving, or transforming relevant rows and columns before answer generation. We first review two main lines of localization-oriented methods: \emph{semantic evidence selection} and \emph{operation-aware table reduction}. We then review existing TableQA benchmarks and evidence annotations, which provide the context for our silver sub-table construction and long-table benchmark.

\paragraph{Evidence localization and retrieval.}
One line of work selects a smaller sub-table containing relevant evidence before reasoning. DATER~\citep{ye2023large} uses LLMs to decompose both table and questions, selecting relevant rows and columns for downstream reasoning. H-STAR~\citep{abhyankar} similarly extracts relevant table regions within a hybrid symbolic--textual framework. Others use retrieval-augmented methods for table reasoning. TableRAG~\citep{tabrag} retrieves relevant schema and cell information for large-scale table understanding, T-RAG~\citep{trag} combines dense table retrieval with generation for open-domain TableQA, and GTR~\citep{gtr} performs graph-based hierarchical retrieval for cross-table question answering. While these methods mainly improve how relevant evidence is retrieved or selected, our approach makes evidence selection operation-aware and treats localization as a question-dependent decision.

\paragraph{Program-guided table decomposition.}
Another direction obtains intermediate tables through explicit operations or executable programs. Chain-of-Table~\citep{wang2024chainoftable} iteratively transforms the table during reasoning and treats intermediate tables as evolving reasoning states. Table-Critic~\citep{YuCW25} iteratively critiques and refines the intermediate table before answer generation. ReAcTable~\citep{ZhangHFCDP24} combines LLM reasoning with tools such as SQL and Python executors to manipulate tabular data step by step. TabSQLify~\citep{TabSQLify} constructs question-relevant sub-tables through SQL queries, while Plan-of-SQLs~\citep{POS} executes a sequence of SQL steps to provide interpretable intermediate reasoning traces. In contrast, \model\ does not require a full operation chain or executable program, and applies decomposition only when localized reasoning is helpful.

\begin{figure*}[t]
    \centering
    \includegraphics[width=\textwidth]{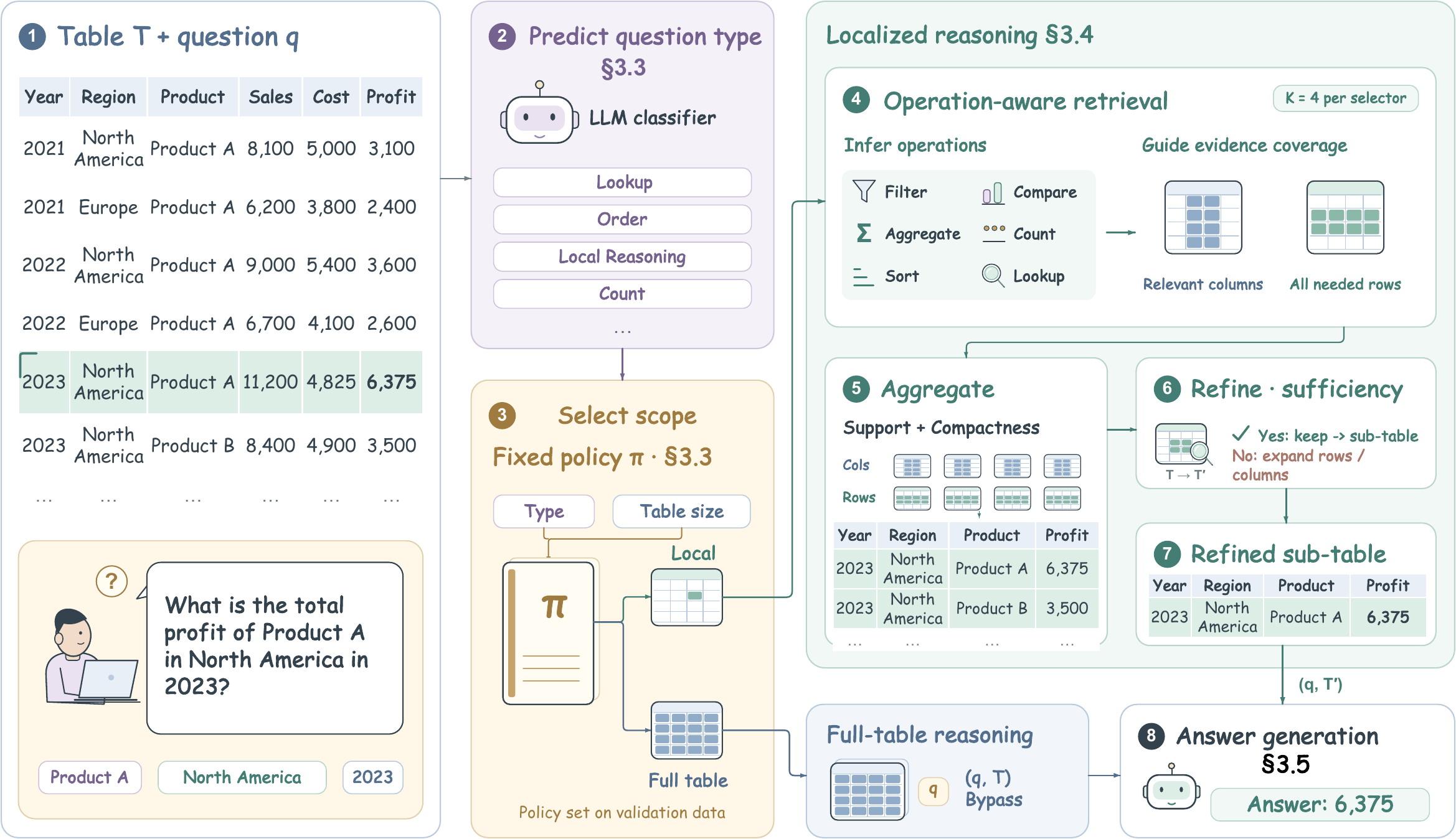}
  \caption{
Overview of \model{}.
The framework selects localized or full-table reasoning for each question. The localized path constructs a refined sub-table, whereas
the full-table path bypasses decomposition.}
    \label{fig:method_overview}
\end{figure*}

\paragraph{Evidence supervision and long-table evaluation.}
Most TableQA benchmarks, such as WikiTableQuestions, provide tables, questions, and final answers, but no explicit annotations of the supporting rows and columns \citep{pasupat}. This limits direct evaluation of intermediate sub-tables. In addition, many commonly used benchmarks contain relatively small tables, typically under 4K table tokens, making them less suited for evaluating long-table reasoning \citep{ChenWCZWLZW20, tablebench, Cheng0WJG0HLZ22}. We address these gaps by constructing silver reference sub-tables for evaluating decomposition quality and by introducing \ldata{}, a benchmark built from real-world long tables.

\section{Methodology}
\label{sec:method}

\subsection{Problem Formulation}

Given a table $T$, a question $q$, and a reference answer $a$, the goal of TableQA is to predict an answer $\hat{a}$ that matches $a$. We represent the table as $T=(H,R)$, where $H={h_1,\ldots,h_m}$ denotes the column headers and $R={r_1,\ldots,r_n}$ denotes the table rows. For each question, a model can either reason over the full table $T$ or reduce it to a question-specific sub-table $T'$. We define the sub-table as $T'=T[R',C']$, where $R'\subseteq R$ and $C'\subseteq H$ are the selected rows and columns. Our goal is not to always minimize the table, but to select the appropriate table scope for each question.

\subsection{Overview of \model{}}

We propose \model{}, a question-adaptive framework that selects the appropriate table scope before answer generation. As illustrated in
Figure~\ref{fig:method_overview}, \model{} consists of three components. (i) An LLM-based scope selector predicts the question type and uses a
fixed policy to choose localized or full-table reasoning. (ii) When localization is selected, an operation-aware decomposer retrieves the
required rows and columns and refines them into a question-specific sub-table. (iii) The answer model generates the answer from the selected input, either the refined sub-table or the original full
table.

\subsection{Question-Adaptive Scope Selection}
\label{sec:scope_selection}

\model{} is designed to determine whether localization should be applied to each question. The scope selector consists of an LLM-based operation classifier $C_{\theta}$ and a validation-calibrated scope policy $\pi$. Given a question $q$ and table $T$, the classifier predicts an operation-defined question type $\hat{\tau}$. The policy then selects the reasoning mode according to the predicted type and table-size regime:

\begin{equation}
\begin{aligned}
\hat{\tau} &= C_{\theta}(q,T), \\
z &= \pi\bigl(\hat{\tau},s(T)\bigr), \\
z &\in \{\mathrm{local},\mathrm{full}\}.
\end{aligned}
\label{eq:scope_selection}
\end{equation}

Here, $s(T)$ denotes the table-size regime. If
$z=\mathrm{local}$, \model{} constructs a question-specific sub-table
before answer generation. Otherwise, it bypasses decomposition and
reasons directly over the original table.

We calibrate $\pi$ through an offline analysis on the validation set.
For each question type, we compare localized reasoning with full-table
reasoning and assign the better-performing mode. We account
for table size when the preferred mode changes on large tables.
At inference time, the LLM predicts only $\hat{\tau}$, while the
policy remains fixed and determines the final scope. The classification prompt is provided in Appendix~\ref{app:scope_selector_prompt}, and the operation definitions and calibration results are reported in Appendix~\ref{app:routing_analysis}.

\subsection{Operation-Aware Table Decomposition}
\label{sec:decomposition}

When the scope selector chooses localized reasoning, \model{} constructs a question-specific sub-table through operation-aware decomposition. The goal is to retrieve the rows and columns needed for the question's reasoning process, rather than relying only on lexical overlap between the question and table content. The decomposer has three steps: operation-aware retrieval, evidence aggregation, and sub-table refinement. All the prompts used are shown in Appendix~\ref{app:prompt_templates}.

\paragraph{Operation-aware retrieval.}
The decomposer first identifies the operation required by the question, such as lookup, filtering, comparison, or counting. This operation determines both the rows and columns to retrieve. For example, for the question \textit{``How many Australian teams scored above 10?''}, the decomposer should retrieve rows whose country is Australia and whose score is greater than 10, together with columns such as country, score, and team identifier. Although the final answer is only a count, the sub-table must preserve the evidence needed to verify the filtering operation. This shows why decomposition should be guided by the operation rather than lexical overlap alone.

\paragraph{Evidence aggregation.}
A single LLM retrieval may be unstable because generation can vary across samples. It may miss necessary evidence or include weakly related rows and columns. To make retrieval more robust, we sample $K$ retrieval outputs for the same question, where $K=4$ by default.

We aggregate row retrievals and column retrievals separately. Each retrieval output is treated as a candidate $c$, which contains either a selected row set or a selected column set. Each candidate is assigned a reliability score $s(c)$. When generation confidence is available for the LLM, $s(c)$ is computed from the average token log probability of the candidate. Otherwise, we set $s(c)=0$, reducing $w(g)$ to frequency-based voting. Candidates that select the same normalized row set or column set are merged into a candidate group $g$.

For each group, we compute a support weight:
\vspace{-0.4cm}
\begin{center}
\small
\begin{equation}
w(g)=\sum_{c\in g}\exp(s(c)).
\end{equation}
\end{center}
Here, $w(g)$ measures how strongly the retrieval samples support group $g$. The term $\exp(s(c))$ converts the candidate score into a positive voting weight, so a group receives higher support when the same evidence set appears multiple times or when its candidates have higher reliability scores.

After aggregation, each distinct row set returned by the retrieval samples forms a row group, and each distinct column set forms a column group. Groups returned more frequently or with higher confidence receive larger weights. For example, if three out of four samples retrieve the same rows, a candidate sub-table containing these rows should receive stronger support.

To construct candidate sub-tables, we first rank the row groups and column groups by their support weights. We progressively union the top-ranked groups to form candidate row sets $R'$ and candidate column sets $C'$, removing duplicate candidates. We then enumerate the resulting row--column combinations and score each candidate sub-table
$T[R',C']$ using the support and compactness criteria defined below. For a candidate row set $R'$, we compute:
\vspace{-0.6cm}
\begin{center}
\small
\begin{equation}
\mathrm{RowScore}(R')=
\frac{
\sum_{g\in G_r;g\subseteq R'} w(g)
}{
\sum_{g\in G_r} w(g)
}.
\end{equation}
\end{center}
Here, $G_r$ is the set of distinct row groups, $g$ is one retrieved row set, and $w(g)$ is its weight. Thus, $\mathrm{RowScore}(R')$ is the proportion of total row-group weight covered by $R'$. The column score $\mathrm{ColScore}(C')$ is computed in the same way over the column groups.

We combine the two scores as:
\vspace{-0.6cm}
\begin{center}
\small
\begin{equation}
\mathrm{Support}(R',C')=
\sqrt{
\mathrm{RowScore}(R')
\cdot
\mathrm{ColScore}(C')
}.
\end{equation}
\end{center}

The support score is high only when the candidate sub-table covers both the rows and columns repeatedly selected across retrieval samples.

We then balance this support with sub-table size:
\vspace{-1cm}
\begin{center}
\small
\begin{equation}
S(R',C')=
\frac{
\mathrm{Support}(R',C')
}{
\rho(R',C')^{\alpha}
},
\rho(R',C')=
\frac{|R'|\cdot|C'|}
{|R|\cdot|H|}.
\end{equation}
\end{center}
Here, $\rho(R',C')$ is the proportion of table cells retained by the candidate, and coefficient $\alpha \in [0.1,0.3]$ controls the penalty on large sub-tables. We select the row--column combination with the highest $S(R',C')$ as the final decomposed table $T'=T[R',C']$.

\paragraph{Sub-table refinement.}
After evidence aggregation, \model{} performs one refinement round by default to verify whether the sub-table $T'$ is sufficient for answering the question $q$. Given $q$, the full table $T$, and $T'$, the verifier determines whether additional rows, columns, or both are needed. If necessary, the missing evidence is added to $T'$, which is then passed to the answer model $M$.

\subsection{Answer Generation}

After scope selection, each question follows a single reasoning path. The QA model $M$ answers from the $(q,T')$ pair when localized reasoning is selected, and from $(q,T)$ pair otherwise.

\section{Experiments}
\label{sec:exp}
We evaluate \model{} to answer three questions. 
\textbf{RQ1:} How effectively does operation-aware decomposition identify compact and sufficient evidence?
\textbf{RQ2:} Can localization mitigate performance degradation on long tables? 
 \textbf{RQ3:} Can question-adaptive scope selection outperform fixed full-table or localized reasoning?

\begin{table}[t]
\centering
\scriptsize
\setlength{\tabcolsep}{2.2pt}
\renewcommand{\arraystretch}{1.08}

\begin{tabular}{@{}lrrrrrr@{}}
\toprule
\multirow{2}{*}{\textbf{Dataset}}
& \multicolumn{3}{c}{\textbf{Table Statistics}}
& \multicolumn{3}{c}{\textbf{\# QA Pairs}} \\
\cmidrule(lr){2-4}
\cmidrule(lr){5-7}
& Cols.
& Rows
& Tokens
& Train
& Valid.
& Test \\
\midrule

\wdata{}
& 6.4
& 25.4
& 662.6
& 11,321
& 2,831
& 4,344 \\

\ldata{}
& 11.0
& 733.8
& 9,786.2
& 1,324
& 239
& 1,110 \\

\subdata{}
& 1.8
& 6.5
& 104.8
& --
& --
& 4,344 \\

\bottomrule
\end{tabular}

\caption{
Statistics of the evaluation resources, averaged over questions. \wdata{} mainly contains small and medium tables, whereas all \ldata{} tables exceed $4{,}096$ serialized tokens. \subdata{} contains silver sub-tables for the $4{,}344$ \wdata{} test questions.
}
\label{tab:dataset_statistics}
\end{table}

\subsection{Experimental Setup}
\paragraph{Base Models.}
We use two base LLMs from different families: LLaMA-3.3-70B~\cite{llama}, and GPT-5-mini~\cite{gpt}. They cover both open-source and closed-source settings, as well as large and relatively smaller LLMs. Detailed settings are provided in Appendix~\ref{app:exp-settings}.

\begin{figure}[t]
     \centering
     \includegraphics[width = 1\linewidth,
        height=0.16\textheight]{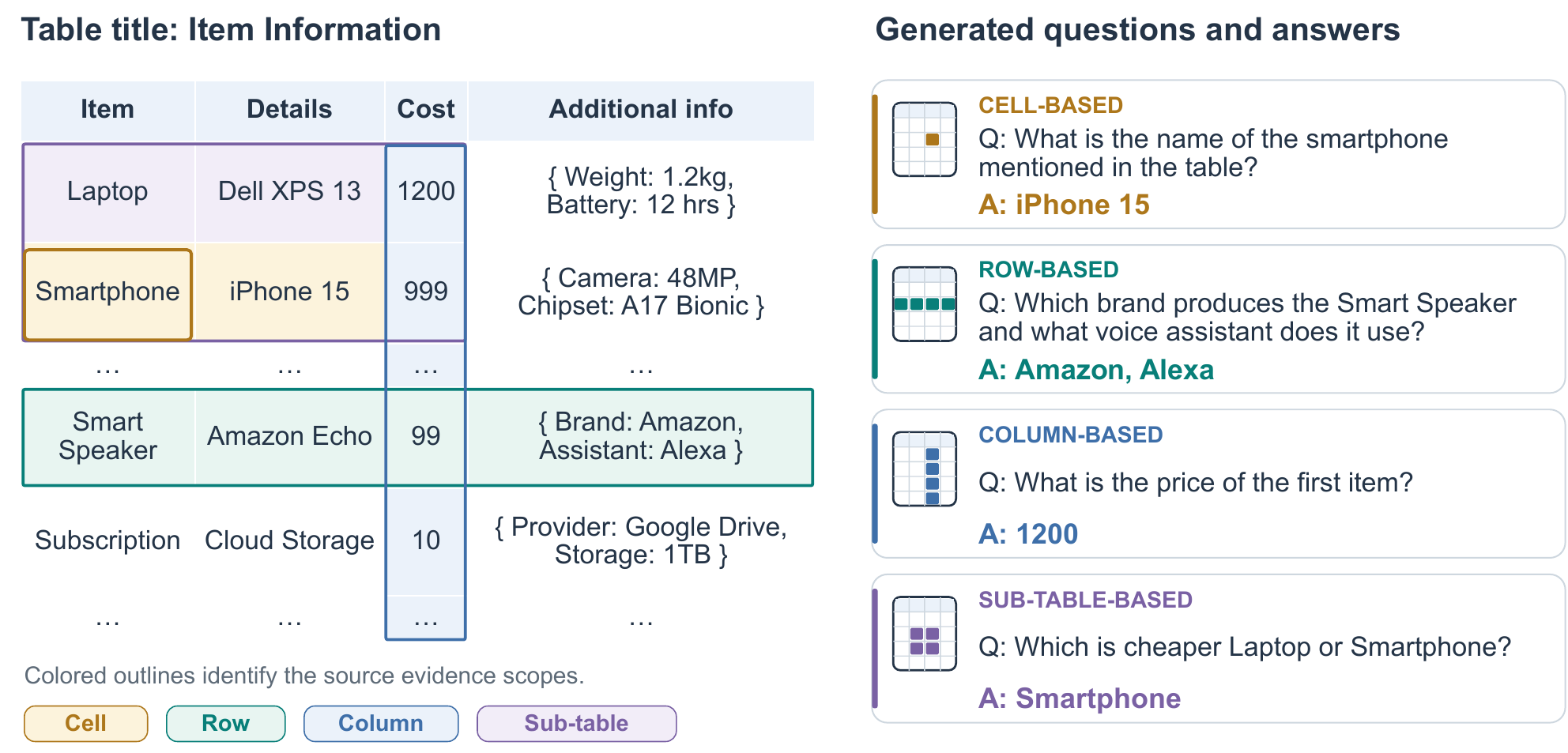}
     \caption{Generated QA pair examples.}
     \label{fig:qa_gen}
\end{figure}

\paragraph{Benchmarks and Evaluation Resources.}
We evaluate \model{} on three benchmarks. Table~\ref{tab:dataset_statistics} shows the statistics for these benchmarks.

\paragraph{WikiTableQuestions (\wdata{}).}
WikiTableQuestions~\citep{pasupat} is a widely used TableQA benchmark
with diverse question types and tables spanning a broad range of
sizes, including tables which sizes exceed
$4{,}096$ tokens. We use \wdata{} for standard TableQA evaluation,
table-size analysis, and type-wise scope analysis.

\begin{table*}[t]
\centering
\small
\renewcommand{\arraystretch}{1.12}
\setlength{\tabcolsep}{4pt}

\begin{tabular*}{\textwidth}{
@{\extracolsep{\fill}}
llcccccc
@{}
}
\toprule
\multirow{2}{*}{\textbf{Category}}
& \multirow{2}{*}{\textbf{Method}}
& \multicolumn{3}{c}{\textbf{\wdata{}}}
& \multicolumn{3}{c}{\textbf{\ldata{}}} \\
\cmidrule(lr){3-5}
\cmidrule(lr){6-8}
&
& \makecell{\textit{GPT}}
& \makecell{\textit{LLaMA}}
& \textbf{Avg.}
& \makecell{\textit{GPT}}
& \makecell{\textit{LLaMA}}
& \textbf{Avg.} \\
\midrule

\multirow[c]{6}{*}{\makecell[c]{Decomposition}}
& TableRAG~\citep{tabrag}
& 69.5
& 56.8
& 63.2
& 47.4
& 49.7
& 48.6 \\

& TabSQLify~\citep{TabSQLify}
& 73.5
& 72.3
& 72.9
& 59.3
& 34.0
& 46.7 \\

& Table-Critic~\citep{YuCW25}
& --
& 70.1
& --
& --
& 52.3
& -- \\

& DATER~\citep{ye2023large}
& 76.3
& 70.5
& 73.4
& 56.0
& 54.2
& 55.1 \\

& Chain-of-Table~\citep{wang2024chainoftable}
& 75.8
& 71.8
& 73.8
& 63.1
& 57.2
& 60.2 \\

\rowcolor{methodgreen}
& \model{} (decomposition-only)
& \underline{82.4}
& 76.8
& 79.6
& \underline{72.3}
& 61.9
& 67.1 \\

\midrule

\multirow[c]{2}{*}{Full-table}
& RoT~\citep{rot}
& 81.7
& \underline{78.7}
& 80.2
& 70.3
& 54.1
& 62.2 \\

& CoT~\cite{wei2022chain}
& 82.1
& 78.5
& \underline{80.3}
& 72.1
& \underline{62.5}
& \underline{67.3} \\

\midrule

\rowcolor{methodgreen}
Adaptive
& \model{}
& \textbf{83.3}
& \textbf{81.1}
& \textbf{82.2}
& \textbf{73.4}
& \textbf{62.8}
& \textbf{68.1} \\

\bottomrule
\end{tabular*}
\caption{
Exact-match accuracy (\%) on \wdata{} and \ldata{}.
The decomposition-only variant applies operation-aware decomposition to every
question, whereas the \model{} adaptively selects between localized and full-table reasoning.
\textit{GPT} and \textit{LLaMA} denote GPT-5-mini and
LLaMA-3.3-70B, respectively. \textit{Avg.} denotes the average across the two answer models and is
omitted when one result is unavailable.
The best and second-best results per column are shown in bold and underlined, respectively. Rows highlighted in green denote variants
of our method.
}
\label{tab:main_results}
\end{table*}

\paragraph{\ldata{}.}
Existing TableQA benchmarks contain relatively few long tables (token size exceeds $4{,}096$ tokens), making
it difficult to systematically evaluate long-table reasoning. We
therefore construct \ldata{}, a dedicated long-table TableQA benchmark.
We extract tables from Spider~\citep{YuZYYWLMLYRZR18} whose table sizes exceed $4{,}096$ tokens, but replace the original
NL-to-SQL questions with newly generated TableQA questions. We use GPT-4o mini~\citep{openai2024gpt4omini} to generate
question--answer pairs from four evidence scopes: a cell, a row, a
column, or a sub-table. The model first identifies an answer supported
by the specified evidence and then formulates the corresponding
question. All generated pairs are manually reviewed with fewer than
$30\%$ requiring correction. The generation prompts and detailed analysis are provided in
Appendix~\ref{app:slqa_construction}. Figure~\ref{fig:qa_gen} shows examples of generated QA pairs.

\paragraph{\subdata{}.}
Final-answer accuracy alone does not reveal whether a decomposition
method retrieves the correct intermediate rows and columns. We
therefore construct \subdata{}, a silver-reference evaluation set
derived from \wdata{}. Each instance consists of a question,
its original table, and a silver sub-table containing the rows and
columns needed to answer or verify the question.

We compare three construction strategies using $200$ randomly sampled
\wdata{} QA pairs with manually annotated gold sub-tables.
Schema-based construction first asks an LLM to identify
question-relevant schema elements and values, which are then mapped back
to the original table. Direct generation instead asks the LLM to
predict the target sub-table directly. The third strategy applies a
verifier to the directly generated sub-table and repairs missing,
invalid, or redundant selections. Direct generation with refinement
achieves the highest agreement with the human annotations and is
therefore used to construct \subdata{}. Detailed results are reported
in Appendix~\ref{app:silver_strategy_selection}.

Given an instance $(q,T,a)$, an LLM selects the minimal supporting rows
and columns. A verifier then detects invalid, incomplete, or
unnecessarily large selections and requests a repair when needed. The
resulting reference sub-table is defined as
$T^*=T[R^*,C^*]$, where $R^*$ and $C^*$ denote the verified rows and
columns.

\subdata{} is used only to evaluate decomposition quality through row-,
column-, and cell-level matching. Its silver annotations are not
available to \model{} during inference. Construction prompts are
provided in Appendix~\ref{app:silver_construction_prompts}.

\paragraph{Baselines.}
We compare \model{} with two groups of baselines.

\textit{Decomposition-based methods} reduce or transform the table
before answering. DATER~\citep{ye2023large} decomposes questions and
tables to select relevant rows and columns. Chain-of-Table~\citep{wang2024chainoftable} iteratively applies table
operations and uses the final transformed table for answer generation. TabSQLify~\citep{TabSQLify} generates SQL queries to construct
question-relevant sub-tables. TableRAG~\citep{tabrag} retrieves relevant
schema and cells for large-table reasoning, while
Table-Critic~\citep{YuCW25} iteratively critiques and refines a
candidate sub-table before answer generation. For a controlled comparison, we use each
decomposition method only to produce its final sub-table. All resulting
sub-tables are passed to the same answer model using an identical
chain-of-thought prompt and answer normalization.

\textit{Full-table reasoning methods} retain the original table during
answer generation. Full-table CoT~\cite{wei2022chain} serializes the complete table and
reasons step by step over it. RoT~\citep{rot} performs row-wise
reasoning before producing the final answer. These baselines represent
two fixed scope strategies: always localizing or always retaining the
full table.

\paragraph{Evaluation Metrics.}
For \wdata{} and \ldata{}, we report answer exact-match accuracy using
the official WikiTableQuestions evaluator~\citep{pasupat}. For
\subdata{}, we independently compare the predicted rows, columns, and
cells with their silver references. We report exact match, which
requires the predicted set to match the reference set completely, and
F1 score, which measures partial overlap between the two sets.

\begin{table*}[t]
\centering
\small
\renewcommand{\arraystretch}{1.12}
\setlength{\tabcolsep}{4pt}

\begin{tabular*}{\textwidth}{
@{\extracolsep{\fill}}
lcccccc
@{}
}
\toprule
\multirow{2}{*}{\textbf{Method}}
& \multicolumn{3}{c}{\textbf{F1}}
& \multicolumn{3}{c}{\textbf{Exact Match}} \\
\cmidrule(lr){2-4}
\cmidrule(lr){5-7}
& Cell
& Row
& Column
& Cell
& Row
& Column \\
\midrule

TableRAG~\citep{tabrag}
& 26.56
& 44.15
& 52.30
& 3.20
& 13.92
& 16.98 \\

TabSQLify~\citep{TabSQLify}
& 40.87
& 56.22
& 75.67
& 9.23
& 26.95
& 39.69 \\

Chain-of-Table~\citep{wang2024chainoftable}
& 52.91
& 76.86
& 67.17
& 25.39
& 56.22
& 38.20 \\

DATER~\citep{ye2023large}
& \underline{68.00}
& \underline{78.46}
& \underline{85.93}
& \underline{36.63}
& \underline{61.50}
& \underline{59.34} \\

\rowcolor{methodgreen}
\model{} (decomposition-only)
& \textbf{70.70}
& \textbf{82.78}
& \textbf{86.95}
& \textbf{41.03}
& \textbf{70.74}
& \textbf{60.54} \\

\bottomrule
\end{tabular*}

\caption{
Sub-table decomposition quality on \subdata{}.
We report F1 and exact match (\%) at the cell, row, and column levels.
Methods are ordered from lower to higher average performance across the
six metrics. The best and second-best results are shown in bold and
underlined, respectively.
}
\label{tab:subtable_results}
\end{table*}

\begin{table}[t]
\centering
\small
\setlength{\tabcolsep}{3.4pt}
\renewcommand{\arraystretch}{1.08}

\begin{tabular}{@{}lccc@{}}
\toprule
Question Type
& \wdata{} $\Delta$
& \ldata{} $\Delta$
& Preferred Scope \\
\midrule
Lookup
& +1.17
& +1.39
& Localized \\

Order/Superlative
& +1.01
& +0.53
& Localized \\

Local Reasoning
& +3.33
& --
& Localized \\

Count-Diff
& +2.13
& --
& Localized \\

Count-General
& -2.40
& -2.34
& Full table \\

Count-Frequency
& -2.68
& +2.22
& Size-dependent \\

Compare
& -1.24
& --
& Full table \\
\bottomrule
\end{tabular}

\caption{
Type-wise localization gain in percentage points.
$\Delta$ is computed as localized reasoning minus full-table reasoning. Positive values favor localization, while negative values favor
full-table reasoning.
\wdata{} is dominated by small and medium tables, whereas \ldata{}
contains only large tables exceeding $4{,}096$ tokens.
``--'' indicates insufficient examples.
Definitions of the question types and the complete type-wise results
are provided in Appendix~\ref{app:routing_analysis}.
}
\label{tab:type_scope_summary}
\end{table}

\subsection{Main Results}
\label{sec:main_results}

Table~\ref{tab:main_results} compares \model{} with fixed full-table
and decomposition-based methods on \wdata{} and \ldata{}, using
GPT-5-mini and LLaMA-3.3-70B as answer models.

\paragraph{RQ1: Operation-aware decomposition.}
The decomposition-only variant consistently outperforms existing
decomposition-based methods. On \wdata{}, it improves over the
strongest baseline by $6.1$ and $4.5$ points with GPT-5-mini and
LLaMA-3.3-70B, respectively. The gains increase to $9.2$ and $4.7$
points on \ldata{}. The larger improvement on \ldata{} suggests that
operation-aware decomposition becomes particularly useful as tables
grow longer and contain more irrelevant rows and columns.

To determine whether these gains arise from better sub-table selection,
we directly evaluate decomposition quality on \subdata{}. Predicted
rows, columns, and cells are compared with the silver references in \subdata{} using
F1 and exact match. F1 measures partial agreement with the reference,
whereas exact match requires the complete predicted set to be correct.

As shown in Table~\ref{tab:subtable_results}, \model{} achieves the best
result across all six metrics. Compared with DATER, the strongest
baseline, it improves cell, row, and column F1 by $2.70$, $4.32$, and
$1.02$ points, respectively. Under the stricter exact-match criterion,
the corresponding gains are $4.40$, $9.24$, and $1.20$ points. The largest improvement occurs in row exact match, indicating that
\model{} more often recovers the complete set of rows required for the
question rather than only a partially relevant subset. At the same
time, the improvements in cell F1 and cell exact match show that the
selected rows and columns are jointly more consistent with the compact
silver sub-tables. The QA gains in
Table~\ref{tab:main_results} are consistent with the row-, column-, and
cell-level results on \subdata{}. Together, these results show that
operation-aware decomposition produces sub-tables that more closely
match the silver references and better support answer generation.

\paragraph{RQ2: Localization on long tables.}
As shown in Table~\ref{tab:main_results}, the decomposition-only
variant achieves an average accuracy of $67.1\%$ on \ldata{},
outperforming the strongest decomposition baseline by $6.9$ points and
nearly matching full-table CoT at $67.3\%$. This is notable because
\ldata{} contains only tables exceeding $4{,}096$ tokens. The result
shows that a large portion of the original table can be removed while
retaining nearly the same overall QA accuracy, suggesting that
localization can mitigate the distracting effect of long-table
context.

However, the similar average performance of localized and full-table
reasoning hides substantial variation across question types.
Table~\ref{tab:type_scope_summary} shows that localization is more
effective when the required rows and columns are concentrated within a
small table region. This includes entity lookup, ranking or extrema,
reasoning over nearby records, and differences between two filtered
counts. In these cases, removing unrelated content makes the required
relationships easier to identify.

Full-table reasoning remains stronger for questions requiring broader
coverage, such as counting all matching rows or comparing multiple
entities or groups. Frequency counting further illustrates the
interaction between question type and table size. It favors full-table
reasoning on \wdata{} but shifts toward localization on \ldata{},
suggesting that preserving the complete table is useful at moderate
sizes, while excessive context becomes increasingly harmful on longer
tables. Localization can therefore reduce long-table degradation, but
its effectiveness depends on both the reasoning requirement and the
amount of table content. Detailed type definitions and complete results
are provided in Appendix~\ref{app:routing_analysis}.

\paragraph{RQ3: Question-adaptive scope selection.}
The complete \model{} augments the decomposition-only variant with scope selection, which determines whether each question should be
answered from the localized sub-table or the original full table. As
shown in Table~\ref{tab:main_results}, this complete framework achieves
the highest accuracy across both datasets and answer models.

Compared with always applying operation-aware decomposition, evidence
scoping improves accuracy by $0.9$ and $4.3$ points on \wdata{} with
GPT-5-mini and LLaMA-3.3-70B, respectively. The corresponding gains on
\ldata{} are $1.1$ and $0.9$ points. These improvements show that even
a strong decomposition method should not be applied uniformly, since
localization may remove rows or columns that remain useful for questions
requiring broader table coverage.

The complete framework also outperforms the strongest full-table
baseline in every setting. On \wdata{}, it improves accuracy by $1.2$
points with GPT-5-mini and $2.4$ points with LLaMA-3.3-70B. On
\ldata{}, the gains are $1.3$ and $0.3$ points, respectively. Although
full-table reasoning retains all table content, it also introduces more
irrelevant rows and columns for questions that depend on a restricted
table region.

These findings are consistent with the type-wise results in
Table~\ref{tab:type_scope_summary}. Questions involving entity lookup,
ranking, extrema, or a small group of related records often benefit
from localization, whereas general counting and multi-entity comparison
typically require broader table coverage. By using scope selection to
select one of these two inputs for each question, \model{} avoids both
unnecessary table reduction and excessive irrelevant context. These results support question-adaptive scope selection. Appendix~\ref{app:scope-selection-analysis} reports oracle diagnostics
and comparisons with direct LLM routing and a learned router.

\paragraph{Ablation results.}
Table~\ref{tab:ablation} shows that all components contribute to the performance of \model{}. Removing evidence aggregation causes the smallest decrease, reducing accuracy by $1.3$ points on \wdata{} and
$0.7$ points on \ldata{}, while removing sub-table refinement leads to slightly larger drops of $1.7$ and $1.5$ points. These results suggest
that aggregation mainly improves the stability of row and column selection, whereas refinement helps recover missing content from initially incomplete sub-tables. Removing operation-aware retrieval has
the largest effect among the decomposition components, decreasing accuracy by $3.1$ points on \wdata{} and $2.0$ points on \ldata{}. This confirms that semantic relevance alone may fail to preserve the rows and columns required for operations such as counting, comparison,
ranking, and aggregation. 

Removing scope selection yields the decomposition-only variant, which applies localization to every question. This reduces accuracy by $4.3$
points on \wdata{} but only $0.9$ points on \ldata{}. The larger drop on \wdata{} is consistent with the type-wise results, where some
questions benefit from retaining the full table. In contrast, the small decrease on \ldata{} shows that operation-aware decomposition remains
effective for most long-table questions by removing substantial irrelevant content while preserving the information needed for answering. Nevertheless, the complete framework still performs best, indicating that adaptive scope selection complements decomposition by retaining the full table when broader coverage is required.

\begin{table}[t]
\centering
\small
\setlength{\tabcolsep}{3.5pt}
\renewcommand{\arraystretch}{1.08}

\begin{tabular*}{\columnwidth}{
@{\extracolsep{\fill}}
lcc
@{}
}
\toprule
\textbf{Variant}
& \textbf{\wdata{}}
& \textbf{\ldata{}} \\
\midrule

\model{}
& \textbf{81.1}
& \textbf{62.8} \\

\quad w/o evidence aggregation
& 79.8 {\scriptsize($-1.3$)}
& 62.1 {\scriptsize($-0.7$)} \\

\quad w/o sub-table refinement
& 79.4 {\scriptsize($-1.7$)}
& 61.3 {\scriptsize($-1.5$)} \\

\quad w/o operation-aware retrieval
& 78.0 {\scriptsize($-3.1$)}
& 60.8 {\scriptsize($-2.0$)} \\

\quad w/o scope selection
& 76.8 {\scriptsize($-4.3$)}
& 61.9 {\scriptsize($-0.9$)} \\

\bottomrule
\end{tabular*}

\caption{
Ablation results using LLaMA-3.3-70B, measured by exact-match accuracy (\%). Values in parentheses denote the decrease relative to the \model{}.
}
\label{tab:ablation}
\end{table}

\section{Conclusion}

We presented \model{}, a framework that treats table scope as a question-dependent decision rather than a fixed preprocessing choice. By combining question-adaptive scope selection with operation-aware
decomposition, \model{} can reduce distracting table content when localization is helpful while preserving the full-table when broader
coverage is required. Experiments on \wdata{} and \ldata{} show that this design improves answer accuracy across both moderate- and large-scale tables, while the results on \subdata{} confirm that the decomposed tables more closely match the rows, columns, and cells needed for answering. These findings suggest that effective TableQA depends
not only on how a table is processed, but also on deciding how much of the table should be retained for each question.

\section{Limitations}

Our evaluation focuses on English TableQA benchmarks and two representative LLMs, with \wdata{} and \ldata{} covering different table scales. Extending the evaluation to additional domains, languages, and model families would provide a broader empirical assessment. Building on our comparisons of type-based, direct LLM, and learned routing, future work could further investigate adaptive scope selection across these broader settings.



\bibliography{custom}
\clearpage
\newpage

\appendix

\section{Appendix}
\label{sec:appendix}
\appendix

\section{Question Type-wise Scope Selection Analysis}
\label{app:routing_analysis}

We derive the scope-selection policy through an offline analysis on the
validation set of \wdata{}. Table~\ref{tab:question_type_definitions}
summarizes the question types definitions.

\begin{table}[t]
\centering
\small
\begin{tabular}{lp{0.62\columnwidth}}
\toprule
Question Type & Definition \\
\midrule
Lookup
& Retrieve an attribute of a specified entity or record. \\

Order/Superlative
& Identify an item through ranking, maximum, minimum, or relative order. \\

Local Reasoning
& Reason over a small, restricted region of the table, such as related or neighboring records. \\

Count-Diff
& Count two filtered subsets and compare or compute the difference between them. \\

Count-General
& Count all records that satisfy one or more conditions. \\

Count-Frequency
& Determine how often a particular value or category occurs. \\

Compare
& Compare values associated with multiple entities or groups. \\
\bottomrule
\end{tabular}
\caption{Definitions of the question types.}
\label{tab:question_type_definitions}
\end{table}

Table~\ref{tab:type_scope_dev} compares localized reasoning with
full-table CoT for each question type. The \textit{Diff.} column is
computed as localized reasoning minus full-table CoT, so a positive
value indicates an advantage for localization. Localized reasoning
performs better for lookup, order/superlative, local reasoning, and
count-difference questions, whose evidence is typically concentrated
within a restricted table region. Full-table reasoning is stronger for
general counting, frequency counting, and comparison questions, which
often require evidence distributed across a broader portion of the
table.

We further examine whether these preferences change with table size.
Table~\ref{tab:type_scope_large} reports the same comparison on the
large-table validation subset. Only question types with sufficient
examples are reported. The overall pattern remains stable, except for
frequency counting, where localized reasoning becomes more effective
on large tables.

Based on these results, we define the scope-selection policy in
Table~\ref{tab:routing_preference}. The default policy follows the
overall validation-set preference for each question type. For large
tables, we adopt the large-table preference when sufficient validation
examples are available. Otherwise, we retain the default choice.

\begin{table}[t]
\centering
\small
\begin{tabular}{lccc}
\toprule
Question Type & Localized & Full-table CoT & Diff. \\
\midrule
Lookup
    & \textbf{86.51\%} & 85.34\% & +1.17 \\
Order/Superlative
    & \textbf{85.01\%} & 84.00\% & +1.01 \\
Local Reasoning
    & \textbf{69.90\%} & 66.67\% & +3.33 \\
Count-Diff
    & \textbf{76.60\%} & 74.47\% & +2.13 \\
Count-General
    & 79.78\% & \textbf{82.18\%} & -2.40 \\
Count-Frequency
    & 82.14\% & \textbf{84.82\%} & -2.68 \\
Compare
    & 67.90\% & \textbf{69.14\%} & -1.24 \\
\bottomrule
\end{tabular}
\caption{
Type-wise accuracy comparison between localized reasoning and full-table CoT
on the validation set of \wdata{}. \textit{Diff.} denotes localized reasoning
minus full-table CoT. The better result for each type is highlighted
in bold.
}
\label{tab:type_scope_dev}
\end{table}

\begin{table}[t]
\centering
\small
\begin{tabular}{lccc}
\toprule
Question Type & Localized & Full-table CoT & Diff. \\
\midrule
Lookup
    & \textbf{70.68\%} & 69.29\% & +1.39 \\
Order/Superlative
    & \textbf{77.66\%} & 77.13\% & +0.53 \\
Count-General
    & 76.83\% & \textbf{79.17\%} & -2.34 \\
Count-Frequency
    & \textbf{84.44\%} & 82.22\% & +2.22 \\
\bottomrule
\end{tabular}
\caption{
Type-wise accuracy comparison on the large-table validation subset of \ldata{}. Question
types with insufficient examples are omitted. \textit{Diff.} denotes
localized reasoning minus full-table CoT.
}
\label{tab:type_scope_large}
\end{table}

\begin{table}[t]
\centering
\small
\begin{tabular}{lcc}
\toprule
Question Type & Default Scope & Large-table Scope \\
\midrule
Lookup
    & Localized & Localized \\
Order/Superlative
    & Localized & Localized \\
Local Reasoning
    & Localized & Localized \\
Count-Diff
    & Localized & Localized \\
Count-General
    & Full table & Full table \\
Count-Frequency
    & Full table & Localized \\
Compare
    & Full table & Full table \\
\bottomrule
\end{tabular}
\caption{
Scope-selection policy derived from
Tables~\ref{tab:type_scope_dev} and~\ref{tab:type_scope_large}.
When the large-table subset does not contain sufficient examples for a
question type, the default validation-set preference is retained.
}
\label{tab:routing_preference}
\end{table}

\section{Prompt Templates}
\label{app:prompt_templates}

This appendix presents the prompt templates used by \model{}.
We omit benchmark-specific demonstrations and implementation metadata
for clarity, while retaining the instructions, input fields, and output
constraints used in our experiments.

\subsection{Operation-Aware Retrieval}
\label{app:retrieval_prompts}

We use separate prompts for row and column retrieval. Both prompts
instruct the model to identify the operation required by the question
and retrieve sufficient evidence for performing and verifying that
operation.

\subsubsection{Column Retrieval}

\begin{promptbox}{Column Retrieval Prompt}
Instruction:
Select the columns needed to answer the question using the
f_col() function.

Return exactly one function call:
f_col([column_1, column_2, ...])

Requirements:
- Copy column names exactly from the table schema.
- First identify the table operation required by the question,
  such as lookup, filtering, comparison, ranking, neighborhood
  retrieval, counting, or aggregation.
- Retain the columns needed to perform and verify the operation,
  rather than selecting columns only by lexical overlap.
- Preserve identifier, condition, comparison, ordering, counting,
  and answer-bearing columns whenever needed.
- Do not output explanations, code, or SQL.

Table:
{LINEARIZED_TABLE}

Optional caption:
{TABLE_CAPTION}

Question:
{QUESTION}

Output:
\end{promptbox}

\subsubsection{Row Retrieval}

\begin{promptbox}{Row Retrieval Prompt}
Instruction:
Select the rows needed to answer the question using the
f_row() function.

Return exactly one function call:
f_row([row 1, row 3, ...])

If all rows are required, return:
f_row([*])

Requirements:
- Use the row numbers shown in the input table.
- First identify the table operation required by the question,
  such as lookup, filtering, comparison, ranking, neighborhood
  retrieval, counting, or aggregation.
- Retrieve enough rows to perform and verify the operation,
  rather than selecting only a likely answer row.
- Preserve comparison sets, anchor rows, neighboring rows, and
  aggregation inputs whenever needed.
- Do not output explanations, code, or SQL.

Table:
{LINEARIZED_TABLE}

Optional caption:
{TABLE_CAPTION}

Question:
{QUESTION}

Output:
\end{promptbox}

\subsection{Sub-table Refinement}
\label{app:refinement_prompt}

\model{} performs one refinement round to
check whether the selected sub-table contains sufficient information
for answering the question. The verifier may retain the current
selection or add missing rows and columns. The refined evidence is
returned as a structured specification and used to construct the final
sub-table.

\begin{promptbox}{Sub-table Refinement Prompt}
You are refining an evidence sub-table for table question
answering.

Given the question, the current sub-table, and the available
information from the original table, determine whether the
sub-table contains sufficient evidence to answer the question.

Return only valid JSON:
{
  "action": "keep" | "expand",
  "final_rows": [row_id, ...],
  "final_columns": ["column_name", ...],
  "reason": "brief explanation"
}

Keep the current rows and columns when the sub-table is
sufficient. Otherwise, add the missing rows, columns, or both.
Use only row IDs and column names provided in the table context.
Do not generate new table values or use external knowledge.

Question:
{QUESTION}

Current sub-table:
{CURRENT_SUBTABLE_WITH_ROW_IDS}

Original table context:
{TABLE_CONTEXT}
\end{promptbox}

\subsection{Answer Generation}
\label{app:qa_prompt}

After scope selection, each question follows only the selected reasoning
path. The placeholder \texttt{\{TABLE\}} is instantiated with the refined
sub-table when localized reasoning is selected, and with the original
table otherwise.

\begin{promptbox}{System Prompt}
You are a precise table question answering assistant.
Follow the required output format exactly.
\end{promptbox}

\begin{promptbox}{User Prompt}
Read the following table and answer the question using only the
information provided in the table.

Table:
{TABLE}

Question:
{QUESTION}

Reason step by step and return exactly two sections:

<think>
Your reasoning
</think>

<answer>
Your final answer
</answer>

Do not include any text outside these two sections.
\end{promptbox}

\subsection{LLM-Based Question-Type Selection Prompt}
\label{app:scope_selector_prompt}

The scope selector does not directly ask the LLM to choose between
localized and full-table reasoning. Instead, the LLM assigns the input
question to one predefined question type. The fixed policy described in
Appendix~\ref{app:routing_analysis} then maps the predicted type and
table-size regime to the final reasoning scope.

\begin{promptbox}{Question-Type Selection Prompt}
Classify the table question according to the reasoning required
to answer it.

Choose exactly one question type:

- "lookup": retrieve a value associated with a specified entity
  or record.
- "order_superlative": identify an item through ranking,
  maximum, minimum, or relative order.
- "local_reasoning": reason over a small set of related or
  neighboring records.
- "count_diff": count two filtered groups and compare or compute
  the difference between them.
- "count_general": count all records satisfying one or more
  conditions.
- "count_frequency": determine how often a value or category
  occurs.
- "compare": compare values associated with multiple entities
  or groups.

Select the type according to the operation and table coverage
required by the question. Do not directly predict whether the
system should use a localized sub-table or the full table.

Question:
{QUESTION}

Table columns:
{TABLE_COLUMNS}

Table size:
{NUMBER_OF_ROWS} rows, {NUMBER_OF_COLUMNS} columns,
{NUMBER_OF_TOKENS} serialized tokens

Table preview:
{TABLE_PREVIEW}

Return only valid JSON:
{
  "question_type": "lookup" | "order_superlative" |
      "local_reasoning" | "count_diff" |
      "count_general" | "count_frequency" | "compare"
}
\end{promptbox}

\section{Detailed Experimental Settings}
\label{app:exp-settings}

\paragraph{Base LLMs and serving.}
For open-source models, we use a local inference backend based on \texttt{swift.llm} with HuggingFace model weights.
For GPT-5-mini, we use the OpenAI Chat Completions API. QA decoding uses a maximum generation budget of 2048 new tokens and model default input budget. Retrieval sampling uses a temperature of $0.5$, while all other components use temperature $0$.


\section{Dataset Construction Details}
\label{app:dataset_construction}

\subsection{Construction of \ldata{}}
\label{app:slqa_construction}

\paragraph{Source tables.}
We construct \ldata{} from existing large tables in
Spider~\citep{YuZYYWLMLYRZR18}. Specifically, we retain tables whose
serialized representations exceed $4{,}096$ tokens under our
preprocessing. We do not reuse the original Spider questions because
they are designed for NL-to-SQL evaluation and closely follow
executable logical forms. Instead, we generate new questions intended
for direct question answering over serialized tables.

\paragraph{Self-adaptive QA generation.}
Our preliminary study found that assigning a predetermined answer to
the model and asking it to generate a corresponding question often
produced unnatural or ambiguous question--answer pairs. The acceptance
rate of this procedure was below $50\%$ in our manual evaluation. We therefore use a self-adaptive generation procedure. Rather than
forcing a predetermined answer, we provide an evidence scope and allow
the model to first identify an answer supported by that evidence and
then generate the corresponding question. We use four evidence scopes:
a single cell, a selected row, a selected column, and a selected
sub-table. These scopes encourage questions requiring different
amounts and configurations of table evidence.

\paragraph{Reasoning coverage.}
We retain both SQL-executable and non-SQL-executable questions.
SQL-executable questions cover structured operations such as
filtering, comparison, counting, and aggregation. Non-SQL-executable
questions support more flexible reasoning over semi-structured table
content and prevent the benchmark from being limited to rigid SQL
logical forms.

\paragraph{Quality control.}
Every generated pair is manually reviewed for answerability, table
grounding, and answer correctness. Valid pairs are accepted directly.
Questions with minor language or ambiguity issues are revised, and
incorrect answers are corrected when the intended question remains
valid. Pairs that cannot be reliably repaired are discarded. More than
$70\%$ of the generated pairs are accepted without modification, while
fewer than $30\%$ require manual correction.

\subsubsection{QA-Pair Generation Prompts}
\label{app:qa_generation_prompts}

We use four prompt variants corresponding to the four evidence scopes.
For cell-based generation, the model selects an answer cell directly.
For the other variants, the selected row, column, or sub-table defines the evidence from which the question and answer are generated.

\begin{promptbox}{Cell-Based QA Generation}
Use the given table as the only source of evidence.

Table header:
{TABLE_HEADER}

Table content:
{TABLE_CONTENT}

Randomly select one cell from the table as the answer and
generate a question whose answer is that cell.

Generate 10 diverse question--answer pairs.

Return one pair per line using exactly the following format:
Q: question; A: answer

Keep each answer concise and return only the answer value.
Do not provide explanations or any text outside the
question--answer pairs.
\end{promptbox}

\begin{promptbox}{Row-Based QA Generation}
Use the given table as the only source of evidence.

Table header:
{TABLE_HEADER}

Table content:
{TABLE_CONTENT}

Selected row:
{SELECTED_ROW}

Generate 4 diverse question--answer pairs using information
supported by the selected row. For each pair, first identify
a valid answer and then formulate a question whose answer can
be derived from the table.

Return one pair per line using exactly the following format:
Q: question; A: answer

The questions must be understandable without referring to
"the selected row" or "the given data." Keep each answer
concise and return only the answer value. Do not provide
explanations.
\end{promptbox}

\begin{promptbox}{Column-Based QA Generation}
Use the given table as the only source of evidence.

Table header:
{TABLE_HEADER}

Table content:
{TABLE_CONTENT}

Selected column:
{SELECTED_COLUMN}

Generate 4 diverse question--answer pairs using information
supported by the selected column. For each pair, first identify
a valid answer and then formulate a question whose answer can
be derived from the table.

Return one pair per line using exactly the following format:
Q: question; A: answer

The questions must be understandable without referring to
"the selected column" or "the given data." Keep each answer
concise and return only the answer value. Do not provide
explanations.
\end{promptbox}

\begin{promptbox}{Sub-Table-Based QA Generation}
Use the given table as the only source of evidence.

Table header:
{TABLE_HEADER}

Table content:
{TABLE_CONTENT}

Selected sub-table:
{SELECTED_SUBTABLE}

Generate 4 diverse question--answer pairs using information
supported by the selected sub-table. For each pair, first
identify a valid answer and then formulate a question whose
answer can be derived from the table.

Return one pair per line using exactly the following format:
Q: question; A: answer

The questions must be understandable without referring to
"the selected sub-table" or "the given data." Keep each answer
concise and return only the answer value. Do not provide
explanations.
\end{promptbox}

\subsection{Selection of the Silver Construction Strategy}
\label{app:silver_strategy_selection}

We randomly sample $150$ WikiTableQuestions QA pairs and manually annotate the gold rows and columns required for each question. We then compare three silver sub-table construction strategies against these
human annotations.

Schema-based construction first generates question-related schema
elements and values and then retrieves the corresponding rows and
columns from the original table. Direct generation predicts the target
sub-table directly from the question, reference answer, and original
table. Direct generation with refinement further uses a verifier to
identify and repair missing, invalid, or redundant selections.

\begin{table*}[t]
\centering
\small
\setlength{\tabcolsep}{5pt}
\renewcommand{\arraystretch}{1.10}

\begin{tabular*}{\textwidth}{
@{\extracolsep{\fill}}
lcccccc
@{}
}
\toprule
\textbf{Construction Method}
& \textbf{Cell F1}
& \textbf{Cell P.}
& \textbf{Cell R.}
& \textbf{Cell EM}
& \textbf{Row F1}
& \textbf{Col. F1} \\
\midrule

Schema-based
& 60.33
& 66.87
& 71.71
& 27.33
& 74.34
& 80.06 \\

Direct Generation
& \underline{71.58}
& \underline{80.59}
& \underline{75.25}
& \underline{42.67}
& \underline{80.38}
& \underline{84.55} \\

Direct Generation + Refinement
& \textbf{74.67}
& \textbf{85.58}
& \textbf{77.48}
& \textbf{44.67}
& \textbf{82.90}
& \textbf{90.15} \\

\bottomrule
\end{tabular*}

\caption{
Comparison of silver sub-table construction strategies on $200$
randomly sampled WikiTableQuestions QA pairs with manually annotated
gold sub-tables. Results are reported as percentages. The best and
second-best results are shown in bold and underlined, respectively.
}
\label{tab:silver_strategy_comparison}
\end{table*}

Table~\ref{tab:silver_strategy_comparison} shows that direct generation
substantially improves over schema-based construction. Verifier-guided
refinement further increases all six metrics, raising cell F1 from
$71.58\%$ to $74.67\%$ and column F1 from $84.55\%$ to $90.15\%$.
We therefore use direct generation with refinement to construct
\subdata{}.

\subsection{Construction of \subdata{}}
\label{app:subtab_construction}

Most TableQA datasets provide only question--table--answer triples and
do not annotate the rows and columns needed to support each answer. We
construct \subdata{} to enable direct evaluation of intermediate
sub-table decomposition.

\paragraph{Direct evidence construction.}
For each \wdata{} instance $(q,T,a)$, an LLM predicts a structured
evidence specification
$E=(\tau,p,R^*,C^*)$. Here, $\tau$ denotes the inferred reasoning type,
$p$ specifies the required evidence scope, and $R^*$ and $C^*$ denote
the selected rows and columns. The reasoning types cover lookup,
filtering, comparison, superlative and ordinal selection, neighborhood
reasoning, counting, aggregation, and Boolean reasoning.

The evidence policy specifies whether the reference should contain only
the answer-bearing evidence, the full comparison domain, all instances
contributing to an aggregation, or the required neighboring context.
The silver sub-table is then extracted from the original table as
$T^*=T[R^*,C^*]$. The model outputs row identifiers and column names
rather than regenerating table contents, ensuring that all selected
evidence is grounded in the original table.

\paragraph{Verifier-guided repair.}
The initial evidence specification may contain invalid indices, omit
necessary evidence, or retain unnecessary table content. A verifier
therefore checks whether the selected rows and columns are valid,
sufficient, and compact. It also examines whether the selected evidence
matches the inferred reasoning type and evidence policy.

When an issue is detected, the verifier feedback is passed to a repair
prompt. The repair step may add missing rows or columns, remove
redundant evidence, or revise the reasoning type and evidence policy.
Only verified constructions are retained as silver references.

\paragraph{Evaluation usage.}
\subdata{} is used to evaluate row selection, column selection, and
cell-level evidence coverage. The reference answers and silver
sub-tables are used only during offline construction and evaluation.
They are not provided to \model{} during inference.

\subsubsection{Silver Sub-Table Construction Prompts}
\label{app:silver_construction_prompts}

We construct silver reference sub-tables using direct evidence
selection followed by verifier-guided repair. Because this procedure
is used only for offline evaluation-set construction, the prompts may
access the reference answers. Implementation-specific fields and
engineering details are omitted for clarity.

\begin{promptbox}{Direct Silver Sub-Table Construction}
Given the question, reference answer, and full table, select the
smallest sub-table containing sufficient evidence to answer or
verify the question.

Return only valid JSON:
{
  "reasoning_type": "lookup" | "filter" | "comparison" |
      "superlative" | "ordinal" | "neighbor" | "count" |
      "aggregation" | "boolean" | "other",
  "evidence_policy": "answer_only" | "comparison_domain" |
      "all_relevant" | "neighbor_context",
  "rows": [row_id, ...],
  "columns": ["column_name", ...],
  "reason": "brief explanation"
}

Use only valid row IDs and exact column names from the table.
Retain all evidence required by the reasoning operation while
avoiding unnecessary rows and columns.

Question:
{QUESTION}

Reference answer:
{GOLD_ANSWERS}

Full table:
{TABLE_MARKDOWN}

Return only the JSON object.
\end{promptbox}

\begin{promptbox}{Silver Sub-Table Repair}
The previously selected sub-table was found to contain missing,
invalid, or unnecessary evidence.

Using the verifier feedback, return the smallest corrected
sub-table that is sufficient to answer or verify the question.

Return only valid JSON:
{
  "reasoning_type": "lookup" | "filter" | "comparison" |
      "superlative" | "ordinal" | "neighbor" | "count" |
      "aggregation" | "boolean" | "other",
  "evidence_policy": "answer_only" | "comparison_domain" |
      "all_relevant" | "neighbor_context",
  "rows": [row_id, ...],
  "columns": ["column_name", ...],
  "reason": "brief explanation"
}

Use only valid row IDs and exact column names from the table.
Add missing evidence and remove unnecessary evidence when
needed.

Question:
{QUESTION}

Reference answer:
{GOLD_ANSWERS}

Previous prediction:
{PREVIOUS_SPEC}

Verifier feedback:
{VERIFIER_FEEDBACK}

Full table:
{TABLE_MARKDOWN}

Return only the corrected JSON object.
\end{promptbox}

\section{Scope Selection Analysis}
\label{app:scope-selection-analysis}

\subsection{Oracle Scope Selection}
The oracle results in Table~\ref{tab:scope-diagnostic} indicate clear room for improvement in scope selection. On \wdata{}, oracle selection improves EM from 82.4\% to 87.8\%, giving a 5.4 point gain over the better fixed branch. On \ldata{}, the corresponding gain is 3.0 points, from 72.3\% to 75.3\%. Since Full and Local achieve similar performance on both datasets, the results suggest that the main benefit comes from selecting the appropriate scope for each instance rather than relying on a single fixed scope.

\begin{table}[t]
\centering
\small
\setlength{\tabcolsep}{4pt}
\begin{tabular}{lrrrrr}
\toprule
Dataset & $N$ & Full & Local & Oracle & Gap \\
\midrule
\wdata{}
& 4,344 & 82.1 & 82.4 & 87.8 & 5.4 \\
\ldata{}
& 1,110 & 72.1 & 72.3 & 75.3 & 3.0 \\
\bottomrule
\end{tabular}
\caption{Full, Local, and Oracle are EM (\%); Gap is oracle EM minus the better fixed-branch EM, in percentage points. Oracle scores are upper bounds for the scope selection.}
\label{tab:scope-diagnostic}
\end{table}

\subsection{Comparison of Scope Selectors}
\label{app:routing-comparison}

\paragraph{Setting.}
We runs use GPT-5-mini as the base model, a QA output budget of 4,096 tokens, and $K=4$ retrieval candidates with refinement. All selectors reuse the same full-table and local answers.

\paragraph{Routing alternatives.}
The fixed type-based selector uses the default policy in Table~\ref{tab:routing_preference}.
Direct LLM routing selects between full-table and localized reasoning
using the question and table information. The learned router uses logistic regression to predict how likely each scope is to produce a correct answer from question and table features. It is trained on $500$ \wdata{} training questions and fixed before test evaluation, selecting the scope with the higher predicted probability. Always-full and always-local use the same scope for every question. The oracle is correct whenever either scope produces a correct answer.

\begin{table}[t]
\centering
\small
\setlength{\tabcolsep}{4pt}
\begin{tabular}{lrrr}
\toprule
Selector & EM & R-Acc  \\
\midrule
Always full
& 82.1 & 42.9  \\
Always local
& 82.4 & 57.1 \\
Type-based (fixed)
& \textbf{83.3} & \textbf{60.0} \\
Direct LLM
& 80.0 & 50.5  \\
Learned router
& 80.5 & 53.3 \\
\midrule
Oracle
& 87.8 & 100.0 \\
\bottomrule
\end{tabular}
\caption{Supplementary routing comparison on \wdata{} test questions.
EM and routing accuracy (R-Acc) are percentages. R-Acc is evaluated
on the questions where branch correctness differs.}
\label{tab:routing-comparison}
\end{table}

\paragraph{Results.}
The fixed type-based policy attains the highest observed EM among
the evaluated routers, exceeding direct LLM routing and the learned
gate by $3.3$ and $2.8$ percentage points, respectively.

\end{document}